\documentclass[runningheads]{llncs}
\usepackage{graphicx}
\usepackage{tikz}
\usepackage{comment}
\usepackage{amsmath,amssymb} % define this before the line numbering.
\usepackage{color}
\usepackage{epsfig}
\usepackage{hyperref}
\usepackage{booktabs}
\usepackage{textcomp}
\usepackage{caption}
\usepackage{url}
\usepackage{hyperref}
\usepackage{colortbl}
\usepackage{multirow}
\usepackage{wrapfig}

\begin{document}
% \linenumbers
\pagestyle{headings}
\mainmatter
\def\ECCVSubNumber{\#6231}  % Insert your submission number here

\title{Unsupervised Monocular Depth Estimation for Night-time Images using Adversarial Domain
Feature Adaptation} % Replace with your title

\begin{comment}
\titlerunning{ECCV-20 submission ID \ECCVSubNumber} 
\authorrunning{ECCV-20 submission ID \ECCVSubNumber} 
\author{Anonymous ECCV submission}
\institute{Paper ID \ECCVSubNumber}
\end{comment}
%******************

%\begin{comment}
\titlerunning{Night time Depth Estimation using ADFA}
\author{Madhu Vankadari\inst{1}%\orcidID{0000-1111-2222-3333}
		 \and
		Sourav Garg\inst{2}%\orcidID{1111-2222-3333-4444}
		 \and
		Anima Majumder\inst{1}%\orcidID{2222--3333-4444-5555}
		\and
		Swagat Kumar\inst{3,1} \thanks{SK will like to thank NVIDIA GPU Grant Program for their support.}%\orcidID{2222--3333-4444-5555}
		\and
		Ardhendu Behera\inst{3}%\orcidID{2222--3333-4444-5555}
		}
\authorrunning{M. Vankadari et al.}
\institute{TATA Consultancy Services, India \\ \email{\{madhu.vankadari,anima.majumder\}@tcs.com}
 \and
Queensland University of Technology, Australia\\ \email{\{s.garg\}@qut.edu.au}\\
%\url{http://www.springer.com/gp/computer-science/lncs}
 \and
Edge Hill University, U.K \\
\email{\{swagat.kumar,beheraa\}@edgehill.ac.uk}
}

%\end{comment}
%******************
\maketitle

\begin{abstract}
In this paper, we look into the problem of estimating per-pixel depth
maps from unconstrained RGB monocular \emph{night-time} images which
is a difficult task  that has not been addressed adequately in the
literature.  The state-of-the-art day-time depth estimation methods
fail miserably when tested with night-time images due to a large
domain shift between them. The usual photometric losses used for
training these networks may not work for night-time images due to the
absence of uniform lighting which is commonly present in day-time
images, making it a difficult problem to solve.  We propose to solve
this problem by posing it as a domain adaptation problem where a
network trained with day-time images is adapted to work for night-time
images.  Specifically, an encoder is trained to generate features from
night-time images that are indistinguishable from those obtained from
day-time images by using a PatchGAN-based adversarial discriminative
learning method.  Unlike the existing methods that directly adapt
depth prediction (network output), we propose to adapt \textit{feature maps}
obtained from the encoder network so that a pre-trained day-time depth
decoder can be directly used for predicting depth from these \textit{adapted
features}.  Hence, the resulting method is termed as ``Adversarial Domain
Feature Adaptation (ADFA)'' and its efficacy is demonstrated through
experimentation on the challenging Oxford night driving dataset. To
the best of our knowledge, this work is a first of its kind to
estimate depth from unconstrained night-time monocular RGB images that
uses a completely unsupervised learning process. The modular
encoder-decoder architecture for the proposed ADFA method allows us to
use the encoder module as a feature extractor which can be used in
many other applications. One such application is demonstrated where
the features obtained from our adapted encoder network are shown to
outperform other state-of-the-art methods in a visual place
recognition problem, thereby, further establishing the usefulness and
effectiveness of the proposed approach.
\end{abstract}

\section{Introduction} \label{sec:intro}
Estimating depth from RGB images is a challenging problem which finds
applications in a wide range of fields such as augmented reality
\cite{pose_ar}, 3D reconstruction \cite{andreas2011stereo},
self-driving cars \cite{handa2014benchmark}, place recognition
\cite{garg2019look}, etc.  The recent success of deep learning methods
has spurred the research in this field leading to the creation of
several new benchmarks that now outperform traditional methods which
rely on handcrafted features and exploit camera geometry and/or camera
motion for depth and pose estimation from monocular or stereo sequence
of images (video).  These learning methods can be broadly classified into two categories: supervised and unsupervised. The supervised  learning methods \cite{eigen2014depth}
\cite{cao2018estimating} necessitate explicit availability of ground truth information (Laser or
LiDAR range data) which may not always be feasible in many real-world
scenarios. This is overcome by the unsupervised methods
\cite{zhou2017unsupervised} \cite{yin2018geonet}
\cite{babu2018undemon} that harness the spatial and/or temporal
consistency present in image sequences to extract the underlying
geometry to be used as the implicit supervision signal required for
training the models.  
\begin{figure}[!t] \centering
  \includegraphics[scale=0.14]{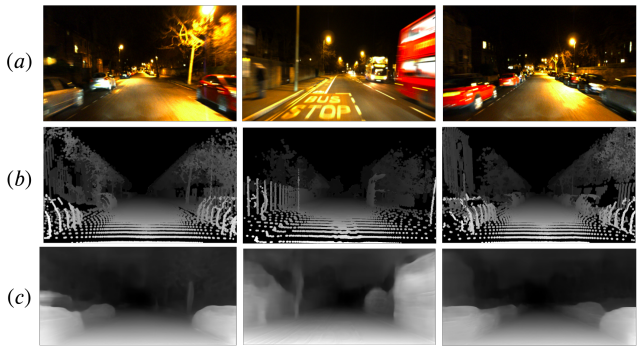} 
  \caption{\small The depth predictions of the proposed method on
  Oxford Night driving images.  Top to bottom: (a) Input RGB
night-time image. (b) Corresponding ground truth depth map generated from the LIDAR
points. (c) The depth predictions using the proposed method }
\label{fig:intro} \end{figure} Many of these methods were shown to
provide very impressive results on several popular datasets such as
KITTI \cite{geiger2013vision} and Cityscapes
\cite{cordts2016cityscapes} containing only day-time images.  In
contrast, there are a very few works that aim to solve the night-time
depth estimation problem, which is comparatively more challenging owing
to factors such as low visibility and non-uniform illumination arising
from multiple (street lights, traffic lights) and possibly, moving light sources (car
headlights). For instance, authors in \cite{im2018robust} exploit
the inherent motion component in burst shot (several successive shots
with varying camera settings, also known as ``auto-bracketing'') to
estimate depth from images taken under low-light condition.
Similarly, Zhu et al. \cite{zhu2019unsupervised} present a deep
learning based method for estimating motion flow, depth and pose from
images obtained from event cameras that return a time-stamped event
tuple whenever a change in pixel intensity is detected. In another
work, Kim et al. \cite{kim2018multispectral} propose a deep network for
estimating depth from thermal images taken during the night time. To
the best of our knowledge, there is no reported work that addresses
the problem of estimating depth and pose directly from a single
ordinary RGB monocular night-time image. The deep learning models
trained on day-time monocular \cite{zhou2017unsupervised} or stereo
images \cite{babu2018undemon} fail miserably on night-time images due
to the inherent large domain shift between these images. The domain
shift refers to the change from day-time conditions (well-lit and
uniform illumination) to night-time conditions comprising low
illumination/visibility with non-uniform illumination caused by
unpredictable appearance and disappearance of multiple point-light
sources (e.g., street lamps or car headlights, etc.).

One possible solution will be to apply image-to-image translation
methods, such as Cycle-GAN \cite{zhu2017unpaired} or MUNIT
\cite{huang2018multimodal}, to map night-time images to day-time
images and then use a pre-trained day-time depth model to estimate
depth from these translated images. Some of these image translation
techniques have been used in the context of night-time images. For
instance, the authors in \cite{anoosheh2019night} use night-to-day
image translation for solving the place recognition problem required
for localization.  Similarly, authors in \cite{atapour2018real}
\cite{zhao2019geometry} explore image translation techniques to generate
synthetic labeled data to reduce the requirement of real-world images
for training depth estimation models. Many of these models trained on
simulated images do not generalize well to natural images due to the
inherent domain shift and hence, employ several domain adaptation
techniques to improve their applicability to real-world situations
\cite{atapour2018real} \cite{zhao2019geometry}
\cite{nath2018adadepth}. These approaches have several limitations.
For instance, many of these methods use two different deep networks -
one for image translation and another for depth estimation, making it
computationally heavy and with possibly, inferior performance due to
the cascading error effect of using two models in a cascade. Since the
image translation module is trained independent of the depth network
module, it may not learn depth-specific attributes required for
preserving structural information during image translation. This may,
in turn, introduce artifacts which might not be understood by the
depth estimation module leading to poor depth prediction for the input
night-time image.  Secondly, it is difficult to generate synthetic
night-time images that can capture all the vagaries of real-world
night conditions as one can observe in the Synthia dataset
\cite{ros2016synthia}. Many of the simulated night-time images in this
dataset appear almost like day-time images and using them for
night-time depth prediction may not give desired results. Finally,
these methods have been applied so far to day-time images for depth
estimation. 

In this paper, we propose a PatchGAN-based domain adaptation technique
for estimating depth from monocular night images by using a single
encoder-decoder type deep network model. Specifically, an encoder
network is trained to generate night-time \textit{features} which are
indistinguishable from those obtained from day-time images. This is achieved by
using an adversarial discriminative learning \cite{tzeng2017adversarial}
that uses day-time encoded features as the reference. These adapted
night features could then be used directly with a decoder network
pre-trained on day-time images for depth estimation. Since the domain
\textit{features} are adapted through adversarial learning, this method is
termed as \emph{``Adversarial Domain Feature Adaptation (ADFA)''}
method to distinguish it from other methods that attempt to adapt
depth predictions directly  \cite{atapour2018real}
\cite{zhao2019geometry} \cite{nath2018adadepth}. Patch{GAN} networks
\cite{isola2017image} \cite{vankadari2019unsupervised} have been shown
to provide superior performance compared to conventional GANs by
capturing high frequency local structural information and hence, form
a natural choice of GAN architecture for the proposed method. 

The resulting outcome of our approach is shown qualitatively in Figure~\ref{fig:intro}. We are able to obtain reliable depth maps shown in
Figure~\ref{fig:intro}(c) from monocular night-time images shown in Figure~\ref{fig:intro}(a). This
is also evident from the interpolated ground-truth depth maps obtained from the
LIDAR point clouds as shown in Figure~\ref{fig:intro}(b). The efficacy of the
proposed approach is demonstrated by applying it to the challenging
Oxford night-time driving dataset \cite{maddern20171}. The modular
encoder-decoder architecture provides the flexibility of using the
encoder module as a feature extractor to extract or select useful
features from input images. Such feature extractors are used in
several applications such as pose estimation \cite{godard2018digging},
Visual Place Recognition (VPR)\cite{garg2020delta}\cite{garg2019look},
object detection \cite{zhao2019object} and segmentation
\cite{badrinarayanan2017segnet}. We demonstrate one such application
where the adapted features obtained from our encoder module are shown
to provide superior place recognition accuracy compared to other
state-of-the-art feature representations available in the literature.

In short, the main contributions made in this paper may be summarized
as follows:

\begin{itemize} \item We propose a novel PatchGAN-based domain \textit{feature}
      adaptation method for estimating depth from unconstrained
      monocular night-time RGB images, which is considered to be more
      difficult compared to day-time images. To the best of our
      knowledge, this is the first instance where adversarial
      discriminative domain feature adaptation is being used for
      estimating depth from unconstrained night-time monocular RGB
      images and this may act as a stepping-stone for future research
      in this field.  

  \item We also propose an image translation-based method for
    night-time depth estimation by using a combination of an image
    translating network (e.g. CycleGAN \cite{zhu2017unpaired}) and a
    standard day-time depth estimation network (such as
    \cite{godard2018digging}) in cascade.  This serves to highlight
    the difficulties involved in such methods and hence, provides
    a strong motivation in favour of the proposed work.

  \item The usefulness and  effectiveness of our method is
    further established by demonstrating that the features obtained
    using the proposed ADFA method outperform other state-of-the-art
    feature representations in a visual place recognition problem.
\end{itemize}

Rest of this paper is organized as follows. The proposed method is
described in the next section. The experimental evaluation of our approach on various datasets is discussed in Section~\ref{sec:experiment}.  The concluding remarks and future scope of this
work is presented in Section~\ref{sec:conc}. 
%Our code will be made available at \url{https://github.com/madhubabuv/NightDepthADFA}

\begin{figure*}[!t] \centering
  \includegraphics[scale=0.15]{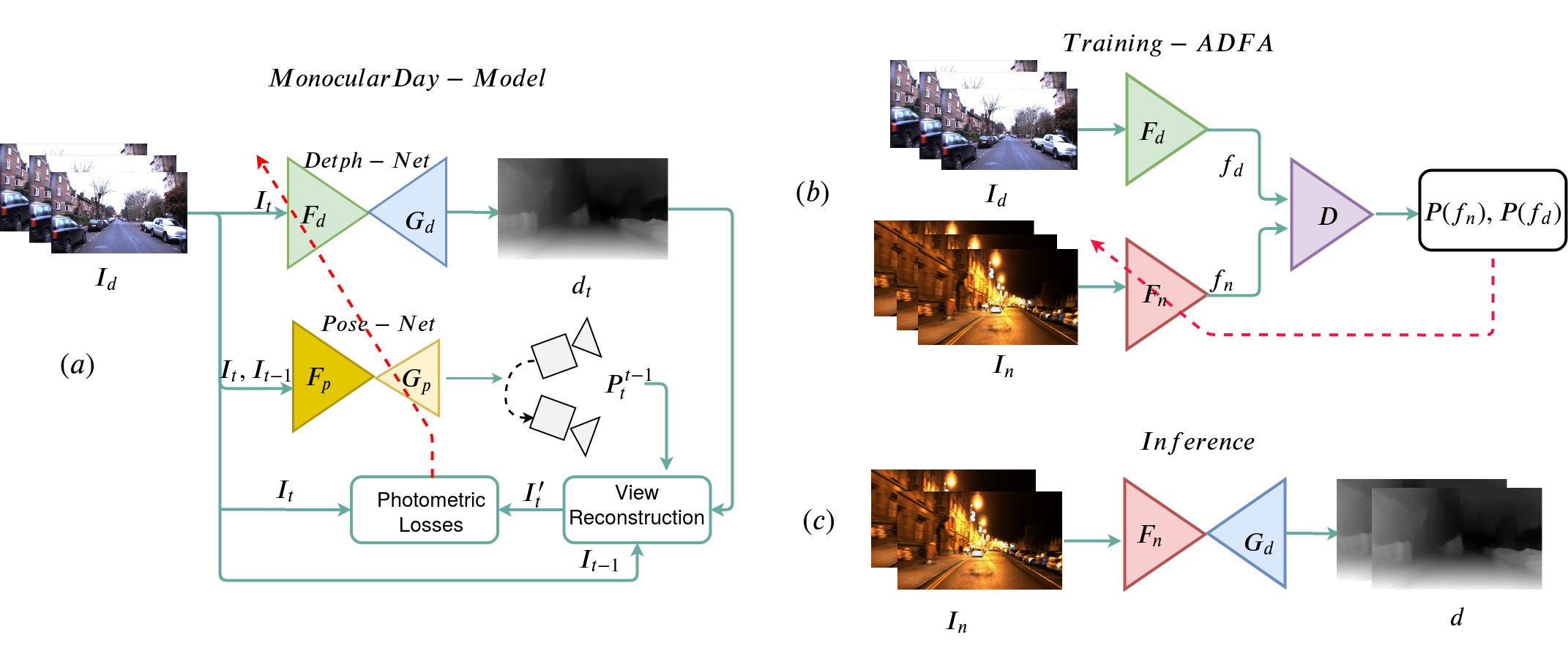} \caption{\small 
  Architectural overview of the proposed method. (a) The monocular
    day-model consists of a Depth-Net ($F_d,G_d$) and a
    Pose-Net ($F_p,G_p$) to predict per-pixel depth-map $d_t$ and 6-DoF
    pose $P^{t-1}_t$, respectively. The day-model is trained using
    photometric losses calculated from images reconstructed from
    view-reconstruction module. (b) A new encoder $F_n$ takes a
    night-time image $I_n$ and predicts $f_n$. The $f_n$ features are
    adapted to look-like day features $f_d$ using adversarial learning
    with a Patch-GAN based discriminator $D$. The red dotted line is drawn to indicate modules trained using back propagation.(c) Finally, the
  new-encoder $F_n$ and the day depth-decoder $G_d$ are used together to
predict depth for night-time images} \label{fig:arch_diag}
\end{figure*}

\section{Proposed method} \label{sec:meth}

We propose to solve the depth estimation problem for night-time images
by posing it as a domain adaption problem in which a model pre-trained on
day-time images is adapted to work for night-time images as well. The
overall approach is shown in Figure \ref{fig:arch_diag}. It consists
of three steps. First, an encoder-decoder type deep network model
($F_d, G_d)$ is trained on day-time images to estimate depth directly
from RGB images by using one of the existing methods as in
\cite{godard2018digging}, \cite{vankadari2019unsupervised}, \cite{yin2018geonet},
\cite{godard2017unsupervised}, \cite{zhou2017unsupervised}.  This is shown in
Figure~\ref{fig:arch_diag}(a). The second step involves training a new
image encoder $F_n$ with night-time images using adversarial
discriminative learning that uses $F_d$ as the generator. This is shown in Figure~\ref{fig:arch_diag}(b). The third and the final step involves using
the new encoder $F_n$ in conjunction with the day-time decoder $G_d$
for estimating depth directly from night-time images as shown in
Figure~\ref{fig:arch_diag}(c). 

The above three components of the proposed ADFA method are
described in detail in the following subsections. 

\subsection{Learning $F_d$ and $G_d$ from day-time images}

Estimating depth from monocular day-time images is an active
field of research where deep learning methods have been applied
successfully and several new benchmarks have been reported in the
literature \cite{eigen2014depth}, \cite{cao2018estimating}, \cite{luo2018single},
\cite{zhou2017unsupervised}, \cite{yin2018geonet}, \cite{vankadari2019unsupervised},
\cite{babu2018undemon}, \cite{luo2018every}. These deep networks have an
encoder-decoder type architecture as shown in Figure~\ref{fig:arch_diag}(a). Such an architecture allows us to decompose
the entire pipeline into two sub-networks, one for encoding (or
extracting) features from input images and another for mapping these
features to depth information. In unsupervised methods, the image
reconstruction error is used as the loss function for training the
entire model thereby avoiding the necessity of having the explicit
ground truth depth information. The images are reconstructed by using
spatial and/or temporal cues obtained from stereo or monocular
sequence of images. The methods that use only temporal cues (such as
optical flow) incorporate an additional network to estimate pose or
ego motion required for image reconstruction
\cite{zhou2017unsupervised},\cite{yin2018geonet}.
The Depth-Net  as shown in Figure~\ref{fig:arch_diag}(a) is composed of
a series of convolutional and deconvolutional layers with different
filter sizes.  Given a monocular day-time image $I_d$, the image
encoder $F_d$ generates, say, $L$ number of convolutional feature maps with
different shapes and sizes, one from each layer. This feature map is
represented as $F_d(I_d)= {f}_d = \{f_d^i\},\ i = 1, 2, \dots, L$,
where $L$ is the total number of convolutional layers used in the
image encoder.  These feature maps are then passed to a
depth-decoder $G_d$ to predict per-pixel depth map $\mathcal{D}$ of
the input image $I_d$.  One can use any of the existing methods
(supervised or unsupervised) to learn the functions $F_d$ and $G_d$.
In this work, we have used the state-of-the art depth-net model
\cite{godard2018digging} as our $F_d$ and $G_d$ which are trained on
the day-time monocular images. Since only monocular sequence of images
are used for training, an additional pose network is required to
estimate ego motion of the camera required for reconstructing images
in the temporal domain.  The encoder network $F_d$ is used to train a
new encoder $F_n$ for night-images using an adversarial learning as
explained in the next section.

\subsection{Learning $F_n$ using night-time images}

Once the day-time image encoder $F_d$ and depth decoder $G_d$ are
learned, our objective is to learn an image encoder $F_n$ that can
generate the features maps $f_n$ from a night-time image $I_n$ which
are indistinguishable from the day-time feature maps $f_d$ obtained
from the day-time encoder $F_d$. There is no direct supervision signal
available for computing the loss function from $f_d$ and $f_n$ as the
input day and night images are \emph{unpaired}. Here, the term
\emph{unpaired} means that these two images are not taken at the same
time or at the same place. The encoder $F_n$ is trained to reduce the
distance between the distributions of day and night feature spaces by
using an adversarial training approach proposed in
\cite{tzeng2017adversarial}. In this approach, the image encoder $F_n$
acts as a \emph{generator} trying to generate feature maps from a
night image $I_n$, which look similar to the day-time feature maps
$f_d$ obtained from a day-time image $I_d$ using a day-time encoder
$F_d$. These generated features maps are then evaluated by a
\emph{discriminator} network $D$ that tries not to get fooled by the
generator by assigning correct labels to them. In this way, the
generator learns to generate day-like feature maps from the
night-time images by playing a zero-sum min-max game with the
discriminator.
	
Unlike a regular GAN discriminator which assigns a single scalar value
for a given input, a patch-based discriminator \cite{isola2017image}
assigns a grid of $m\times n$ scalar values for a given feature map.
Each value of this grid is a probability ranging from $0$ (night) to
$1$ (day) and it corresponds to a patch of the input feature map. This
allows the discriminator to evaluate the input feature maps locally
thereby, providing superior distinguishing ability compared to normal
GAN discriminators. In addition, the patch-based discriminators
are fully convolutional and hence, are computationally much faster
compared to the other discriminator models that use fully-connected layers along with the convolutional layers~\cite{vankadari2019unsupervised}.  
	
Instead of training a single discriminator network on the feature maps
obtained from the final convolutional layer of the image encoder as is
done in \cite{nath2018adadepth}\cite{tzeng2017adversarial}, we train
multiple discriminators, one for each layer of the encoder network to
constrain the solution space further. Hence, the proposed multi-stage
patch-based discriminator is composed of $L$ number of discriminators
where each discriminator $D_i,$ takes feature maps $(f_n^i,f_d^i)$ obtained from the $i-$th convolutional layer of the encoder networks $(F_n, F_d)$ as input. This multi-stage discriminator is shown to provide superior
domain adaptation performance which will be discussed later in the experiments
section.

\subsection{Training Losses}
The proposed method is an unsupervised learning approach which neither uses any explicit ground truth nor paired day-night image examples
to calculate losses for training. Instead, we entirely rely on
adversarial losses calculated using the discriminator module. The loss
functions to learn $F_n$ and $D$ can be expressed as follows:
\begin{eqnarray} 
\label{eqn:gan_loss}
 \mathcal{L}_{GAN}(F_n,D) &  =   &\min_{F_n}\max_{D}V(F_n,D)= \mathbb{E}_{f_d \sim F_d(I_d)}[\log(D(f_d))]\:\nonumber \\
 						 & &\qquad \qquad +\mathbb{E}_{f_n \sim F_n(I_n)}\left[\log(1-D(f_n))\right] \\  
  				\min_{F_n} L_{{F_n}}(F_n,D,I_n) & =  & \frac{1}{L}\sum_{i=1}^{L} -\:\mathbb{E}_{f_n \sim F_n(I_n)}\left[\sum_{m,n}\log\left[D_i(f_n^i)\right]_{m,n}\right] \\ 
 				\min_{D} L_{D}(F_d,F_n,D,I_d,I_n)& = & \frac{1}{L}\sum_{i=1}^{L}-\:\mathbb{E}_{f_d \sim F_d(I_d)}\left[\sum_{m,n}\log\left[D_i(f_d^i)\right]_{m,n}\right] \:\nonumber \\
 				& &\quad -\:\mathbb{E}_{f_n \sim F_n(I_n)}\left[\sum_{m,n}\log\left(1-\left[D_i(f_n^i)\right]_{m,n}\right)\right]
 \end{eqnarray}
 The details about our experimental setup and various experiments conducted are explained in the following section.

\section{Experiments and Results} \label{sec:experiment}

 In this section, we provide various experimental results to establish
 the efficacy of the proposed method for estimating depth from
 night-time monocular RGB images. We use the publicly available Oxford Robotcar
 dataset \cite{maddern20171} for evaluating the performance of our
 method. This dataset is used to perform two sets of experiments. The
 first experiment is carried out to analyze the depth estimation
 performance of the proposed method while the second experiment is
 performed to demonstrate the flexibility of using the encoder for
 solving a Visual Place Recognition (VPR) problem. The overview of
 dataset used and the details of experiments performed are described
 next in this section.

\subsection{Oxford Robotcar Dataset : Training and Testing Data Setup }
Oxford RobotCar dataset \cite{maddern20171} is a popular outdoor-driving
dataset comprising of images
collected during different seasons, weather conditions and at
different timings of day and night. The data collection is carried out
over a period of one year by setting cameras in all the four
directions. The images captured from the front-view stereo cameras are of resolution $1280
\times 960$. We have used the left images of the front stereo-camera
(Bumblebee XB3) data from the sequences captured on 2014-12-16-18-44-24
for night-time and 2014-12-09-13-21-02 for day-time images for depth
estimation. The training is performed on the images from the first 5
splits of the day and night-time sequences after cropping the car-hood
from the images and downscaling them to $256\times512$. The static
images where the car has stopped at signals are not considered for the
experiments and thus, the total number of images left for training is close to 20,000.
We have randomly sampled a total of 498 images for testing from
the 6th split of night-driving sequence.

For VPR, we have used day and night sequences as 2014-12-09-13-21-02
and 2014-12-10-18-10-50 respectively from the Oxford Robotcar dataset, where the query (night) sequence is
different from that used in the network training. We only used the
first $6000$ \textit{stereo-left} image frames from each of these
traverses which were uniformly sub-sampled using the GPS data to
maintain consecutive frame distance of approximately $2$ meters. The
day traverse is used as the reference traverse against which each of
the query (night) image representations is compared with Euclidean distance to
retrieve the best match. The night images do not overlap geographically with the night data used for training
the model employed for feature extraction for VPR experiments. The evaluation is done using the GPS data by calculating
the recall rate for the localization radius varying between $0$ to
$100$ meters. Here, recall rate is defined as the ratio of correctly
matched images within the given radius of localization to the total
number of query images.

 \begin{figure*}[!t] 
    \centering
    \includegraphics[scale=0.21]{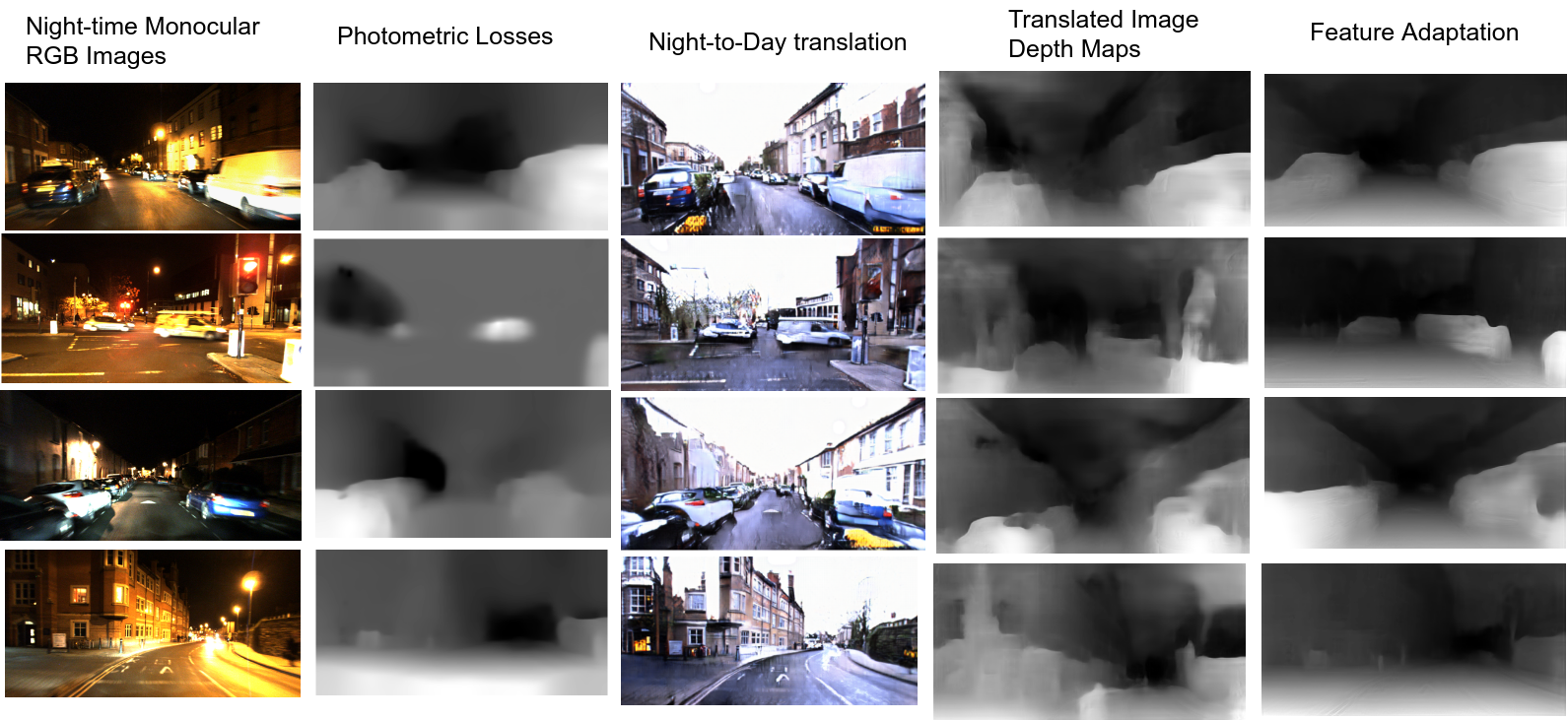}
    \caption{\small A qualitative comparison of predicted depth-maps with
    different experiments. The first column shows the night-time images which are provided as input to different networks. The second column shows the output depth images obtained using photometric losses. As one can observe, these methods fail to maintain the structural layout of the scene. The third column shows the output of an image-translation network (CycleGAN) which are then applied to a day-depth estimation network to obtain depth-maps as  shown in the fourth column. These are slightly better compared to the previous case but it introduces several artifacts which degrade the depth estimation in several cases. The last column shows the predictions using the proposed ADFA approach. As one can see, the proposed method provides better predictions compared to these methods and is capable of preserving structural attributes of the scene to a greater extent} \label{fig:quality_comp} 
\end{figure*}

\subsection{Experimental setup}

The proposed method is implemented using TensorFlow
\cite{abadi2016tensorflow}. The network is trained for 40 epochs
using a GTX 1080 Alienware-R4 laptop. The learning rate is  initially
set to 0.0001, then it is reduced by half after $3/5^{th}$ of the
total iterations and finally, it is further reduced by half after
$4/5^{th}$ of the total iterations. Leaky Relu \cite{xu2015empirical}
is used as an activation function in all the layers, except in
disparity prediction layers. The predicted disparity is normalized to
have the maximum disparity as 30 percent of the input image width by
using sigmoid as activation function while learning the day-time depth estimation model. 
%$\lambda$ value in Equation \ref{eqn:smoothness_loss} is set to
%$0.1/s$, where s is the scale of the disparity-map with respect to the
%input image resolution.
The network is trained using the Adam
\cite{kingma2014adam} optimizer. Two major experimental studies, one
for depth estimation and another for visual place recognition, are
carried out under extreme photometric variations using the Oxford
dataset~\cite{maddern20171}. Both the experimental studies along with
the qualitative and quantitative analyses are presented
below. 

%\begin{table}[!tphb]
%\centering
%\caption{Comparing two approaches of depth estimation in terms of
%model size and training time.}
%\label{tab:model_comp}
%\footnotesize{
%\renewcommand{\arraystretch}{1.3}
%\begin{tabular}{|p{2.5cm}|p{2cm}|p{2cm}| }
% \hline 
%   & Cycle-GAN + DepthNet & ADFA \\
%  \hline
%  Training Time & 60 hours & 8 hours\\ \hline
%  \# Tunable Parameters  & 19M   & 8M \\
% \hline
% \end{tabular}}
% \end{table}

\begin{table*}[!t]
\centering
\caption{ \small A quantitative performance comparison analysis for depth
  estimation from night-time images. The top split of the table is
  evaluated with 60 meters and the lower is evaluate with 40 meters as
  the maximum depth-range. Higher value is better for the blue color
labeled cells and lower value is better for the rest}
\label{tab:oxford_results}  
\scriptsize{
\renewcommand{\arraystretch}{1.5}
\centering
\begin{tabular*}{\textwidth}{|c @{\extracolsep{\fill}}|c|c|c|c|c|c|c|}
  %{ |p{3cm}|p{0.5cm}|p{1.0cm}|p{1.0cm}|p{1.0cm}|p{1cm}|p{1cm}|p{1cm}|p{1cm}|p{1cm}|p{1cm}| }
 \hline
 \multirow{2}{*}{Method}   & \multicolumn{4}{c|}{Error Metric $\downarrow$} & \multicolumn{3}{c|}{Accuracy Metric $\uparrow$} \\ \cline{2-8} 

   &    Abs Rel & Sq Rel & RMSE & logRMSE & \cellcolor[HTML]{7BABF7} $\delta <$1.25
  &\cellcolor[HTML]{7BABF7}$\delta <$1.25$^2$ & \cellcolor[HTML]{7BABF7}$\delta <$1.25$^3$ \\ 
 \hline

 Monodepth2 \cite{godard2018digging} (Day) & 0.7221 &    11.5155 &   14.253 &    0.663 &   0.252 &  0.467 &    0.644 \\
 
 Monodepth2 \cite{godard2018digging} (Night) &  0.3990 &   38.8965 &    23.596 &     0.408 &  0.482 &   0.760  & 0.894 \\
 
 Cycle-GAN \cite{zhu2017unpaired}   & 0.7587 & 12.7944 &   13.681 & 0.663 & 0.277 &      0.503 &   0.688	 \\
 
 ADFA (with KITTI)					&      0.3589 & 5.1174 &     11.611 &   0.384 &  0.424 &   0.730  &   0.914	 \\
 
 ADFA (with Oxford)        			&    \textbf{0.2327} &   \textbf{3.783} &     \textbf{10.089} &     \textbf{0.319} &     \textbf{0.668} &    \textbf{0.844} &     \textbf{0.924} \\
 
% Ours (kitti+oxford)			&    0.2338 &  3.9668 & 10.169 &   0.314 &   0.675 &    0.841 &    0.922 \\
 
 \hline 
 
 Monodepth2 \cite{godard2018digging} (Day)&  0.6108 &     6.9513 &     9.945  &     0.592 &   0.267 &    0.502 &      0.695\\
 
 Monodepth2 \cite{godard2018digging} (Night)&  0.2921&     7.5395&     10.686 &     0.332    &   0.588 &     0.829 &   0.932 \\
 
 Cycle-GAN \cite{zhu2017unpaired}   & 0.6497 &      7.9346 &      9.521 &    0.596 &  0.298 &      0.546&      0.740\\
 
 ADFA (with KITTI)					 & 0.2984 &   3.2349 &      7.801 &    0.328 &  0.495 &   0.833 &    0.942	 \\
 
 ADFA (with Oxford)        			 &  \textbf{0.2005}&     \textbf{2.5750} &      \textbf{7.172} &      \textbf{0.278} &  \textbf{0.735} &   \textbf{0.883} & \textbf{0.942} \\
 
% Ours (kitti+oxford)			&    0.2003 &     2.6959  &      7.224 &     0.275 & 0.746 &     0.877 & 0.940 \\
 
 \hline
\end{tabular*}}

\end{table*}

\subsection{Study 1: Depth Evaluation}

In this study, we perform several experiments to establish the
efficacy of our proposed method for estimating depth from monocular
night-time images. The summary of these experiments is provided in
Table~\ref{tab:oxford_results}. The first row of this table shows the
outcome of our first experiment where we train a monocular version of
Monodepth2 \cite{godard2018digging} network on Oxford day-time images
and then, test it on Oxford night-time images. As expected, the
day-time trained model performs poorly on night-time images because of
the inherent domain shift present between day-time and night-time
images. The second row shows the outcome of another experiment where
the same network is trained on the Oxford night-time images and then,
tested on a different set of night-time images (test-split). The
performance in this case is better than the first experiment but still
not good enough as the presence of temporal intensity gradient makes
it difficult to use the existing photometric losses for training the
network. The third row of this table shows the outcome of yet another
experiment where we use image translation for depth estimation.
In this approach, we use Cycle-GAN \cite{zhu2017unpaired} for
translating night-time Oxford images into day-time images and then use
a day-time trained Monodepth2 model for estimating depth from these
translated images. The performance of this approach is similar to the
above methods (worse in terms of `Abs Rel' metric and better in terms
of `RMSE' metric) indicating that image translation is not adequate
for solving the night-time depth estimation problem. Moreover, it is a
computationally expensive method that uses two independent networks in
cascade unlike the above methods that use only one network for this
task. We now apply our proposed ADFA method to adapt the depth model
used in the first experiment above and the outcome is shown in the
fifth row of this table. As one can see, it provides significant
improvement over the previous three approaches, thereby establishing
the superiority of our approach. In this case, day-time
encoder-decoder pair ($F_d$, $G_d$) and night-time encoder  ($F_n$) are
trained using images from Oxford dataset and then tested using
night-time images from the same dataset. We also perform another
experiment where the day-time encoder-decoder network ($F_d$, $G_d$) is
trained on the KITTI dataset, but the night-time encoder ($F_n$) is
trained and then tested on night-time Oxford images. The corresponding
result is shown in the fourth row and is labeled as `ADFA (with
KITTI)'. While its performance is worse than ADFA (Oxford), it is better than
all other methods mentioned above. It is worth to mention that this is an extreme
case of domain adaption where not only there is a domain variation
from day to night, but also a place variation from KITTI to Oxford. It
only demonstrates the resilience of our approach whose performance
degrades gracefully in the face of this extreme domain variation. Even
though Monodepth2 \cite{godard2018digging} model has been used as our
base network architecture for providing the above performance
analysis, ADFA is a generic approach which could be applied to any
other deep network model with similar effect.

\begin{table*}[!t]
\caption{\small Ablation study to determine the number of day-encoder
convolutional layers to be used during the adversarial learning. The
best performance is achieved by skipping the first two layers (without
cnv-1,2) features}  
\label{tab:ablation_study}
\centering
{\scriptsize 
\renewcommand{\arraystretch}{1.5}
\begin{tabular}{ |p{3cm}|p{1.5cm}|p{1.5cm}|p{1.5cm}|p{1.8cm}|  }
 \hline
  Method & Abs Rel & Sq Rel & RMSE & logRMSE \\
 \hline

 Full conv layers  &  0.2071&     2.8971 &      7.619 &      0.282 \\ 
 
 without cnv-1 & 0.2038 &  2.7908 &     7.461 &     0.280\\
  
 without cnv-1,2 & \textbf{0.2005}& \textbf{2.575}& \textbf{7.172} & \textbf{0.278} \\

 without cnv-1,2,3 & 0.2260 & 2.574 & 7.283 &0.300  \\
  \hline
\end{tabular}}
\end{table*}

%Unlike Monodepth2
%\cite{godard2018digging}, our method does not require photometric
%losses for training the model and hence, does not suffer from the
%problem arising out of temporal intensity gradient present in the
%image sequences. 
%Since Monodepth2 is a
%state-of-the-art method for estimating depth from monocular images and
%hence, it has been used here as a baseline approach for evaluating the
%performance of the proposed approach. However, our proposed domain
%feature adaptation method could be applied to any other depth network
%model to obtain similar results. 

A qualitative performance comparison
of these methods is shown in Figure~\ref{fig:quality_comp}. The first
column shows the input night-time images selected randomly from the
test set. The second column shows the depth estimation results
obtained by using methods such as Monodepth2 \cite{godard2018digging}
that use photometric losses for training. The third column shows the
images obtained after image translation by using methods such as
Cycle-GAN \cite{zhu2017unpaired}. The fourth column shows the depth
map obtained from these translated images by a pre-trained day-time
depth network.  We can clearly see that image translation introduces
several artifacts leading to poor depth estimation results. The last
column shows the depth prediction results obtained by using our
proposed ADFA method.  One can clearly notice the improvements
achieved through our proposed domain feature adaptation method.

The front LMS laser sensor data with INS data is used to prepare the
ground-truth needed for testing images using the official code-base
released with the dataset. The maximum depth range is set to 60m in
the first half of the Table~\ref{tab:oxford_results} and changed to
40m in the second half. The scale is calculated using the ground-truth
depth data, as it is done in
\cite{zhou2017unsupervised,yin2018geonet}. 
%To the best of our
%knowledge, the approach of estimating depth from night-time image is a
%first of its kind and there exits no other deep learning based
%approach in the literature for the works in this direction. 
In addition, an ablation study is carried out to determine the optimal number of night-time encoder $F_n$ layers to be constrained for the best performance and the results are shown
in Table~\ref{tab:ablation_study}. We observed that a model trained by
skipping the first two layers of the day-encoder gives the
best-performance and the same model is used to report the final
results.

\begin{figure}[!t]
    \centering
    \includegraphics[scale=0.3]{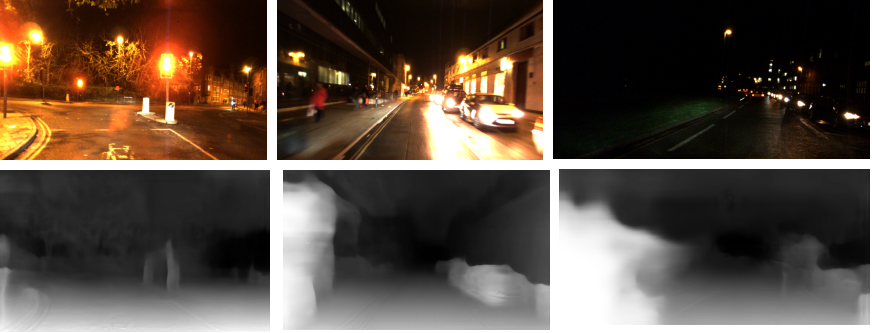}
    \caption{\small Failure cases of the proposed depth prediction approach.
    The model is not able to predict accurate depth for blurred image
  regions, traffic signal poles and very low illuminated regions of
the image }
    \label{fig:failure_cases}
\end{figure}

To the best of our knowledge, the proposed work is the first attempt at solving the depth estimation problem
for unconstrained night-time monocular images for which no priors are available in the literature. However, there are some cases, shown in the  Figure~\ref{fig:failure_cases}, where the model is observed to provide poor 
or failed prediction results. Some of the failure cases include
night-time images with very low-illumination conditions, blurred image regions and saturated regions (bright light spots). It is also difficult for our method to deal with small and narrow structures such as traffic poles. The
failure case with low-illuminated night-time images could be due to
the absence of such extreme conditions in day-time images on which the
day encoder-decoder model is trained. The problems associated with
small structures could be dealt by incorporating some semantic
information (if available) into the training data. These limitations
will provide a fertile ground for further research in this field. 

%	 \begin{figure}
%  \centering
%  \includegraphics[scale=0.4]{../figs/youtube_comp.png}
%  \caption{Disparity prediction results of images randomly taken from a YouTube video \cite{naweed2019driving} when directly tested  with the proposed model trained on Oxford Day \& Night images.}
%  \label{fig:youtube_comparisono}
%\end{figure}

\begin{figure*}[!t]
    \centering
    \includegraphics[scale=0.4]{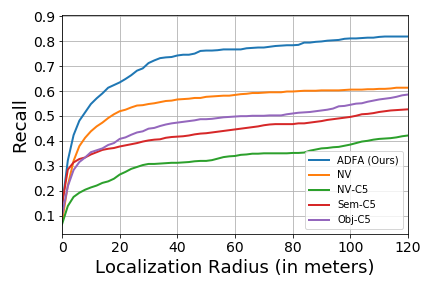}
    \caption{\small Visual Place Recognition Performance Benchmark: It can be observed that the feature representations derived from our depth encoder perform the best as compared to other approaches}
    \label{fig:vpr_curve}
\end{figure*}

\newcommand{\scaleOne}{0.12}
\newcommand{\vsOne}{0.71cm}
\begin{figure*}[t]
\centering

% \begin{minipage}[b]{\linewidth}
% \centering
%   \includegraphics[scale=0.09]{qualImages/qualRes_q110.png}
% \end{minipage}\\
% \begin{minipage}[b]{\linewidth}
% \centering
%   \includegraphics[scale=0.09]{qualImages/qualRes_q372.png}
% \end{minipage}\\
% \begin{minipage}[b]{\linewidth}
% \centering
%   \includegraphics[scale=0.09]{qualImages/qualRes_q558.png}
% \end{minipage}\\
% \begin{minipage}[b]{\linewidth}
% \centering
%   \includegraphics[scale=0.09]{qualImages/qualRes_q608.png}
%   \centerline{\hfill Query \hfill \ \ \ \ GT \hfill \ \ Ours-C5 \hfill NV \hfill Obj-C5 \hfill Sem-C5 \hfill NV-C5 \hfill}
% \end{minipage}

\begin{tabular}{rcccc}
\centering
  Query \vspace{\vsOne} & \multirow{7}{*}{\includegraphics[scale=\scaleOne]{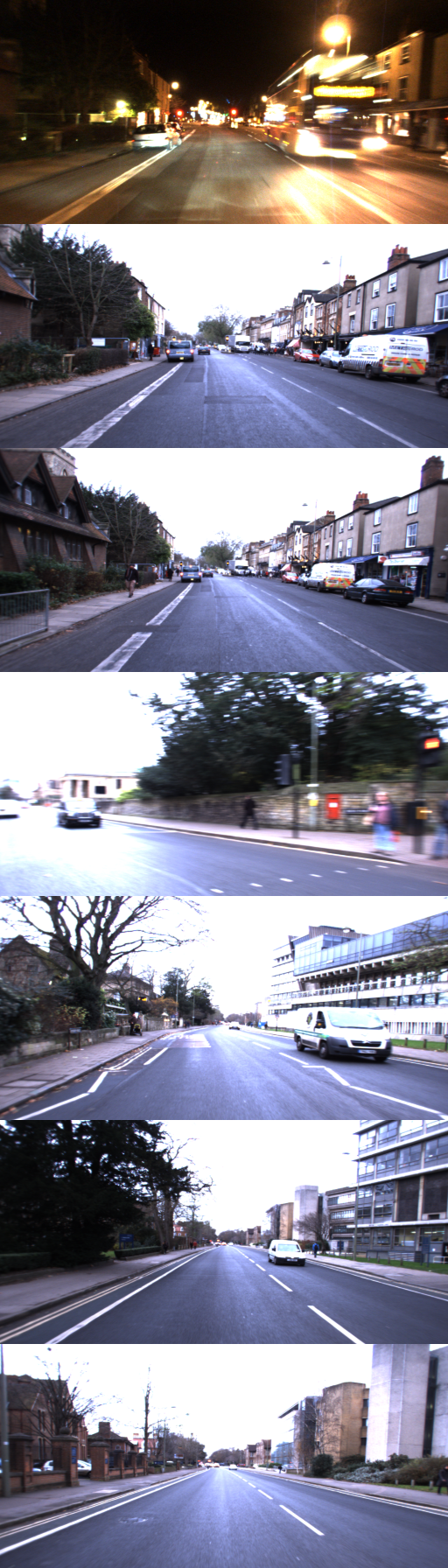}} &
  \multirow{7}{*}{\includegraphics[scale=\scaleOne]{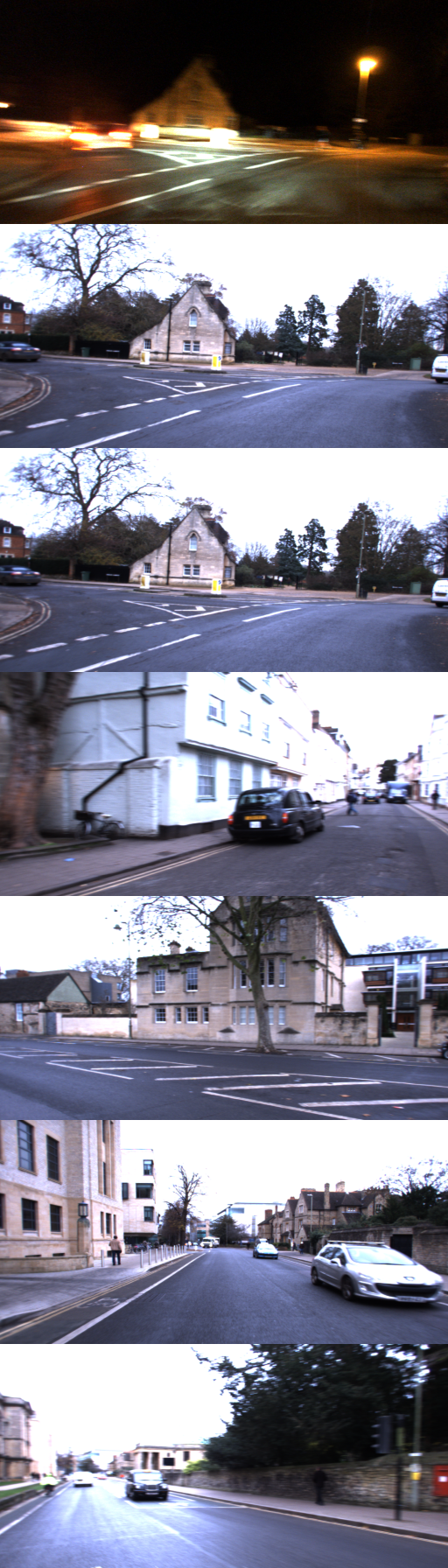}} &
  \multirow{7}{*}{\includegraphics[scale=\scaleOne]{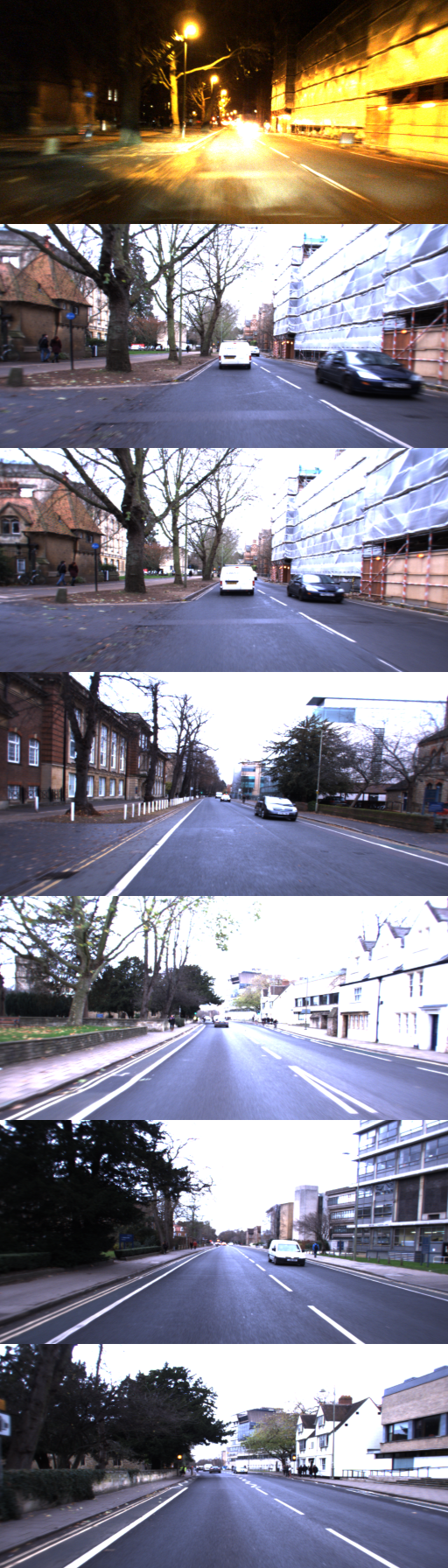}} &
  \multirow{7}{*}{\includegraphics[scale=\scaleOne]{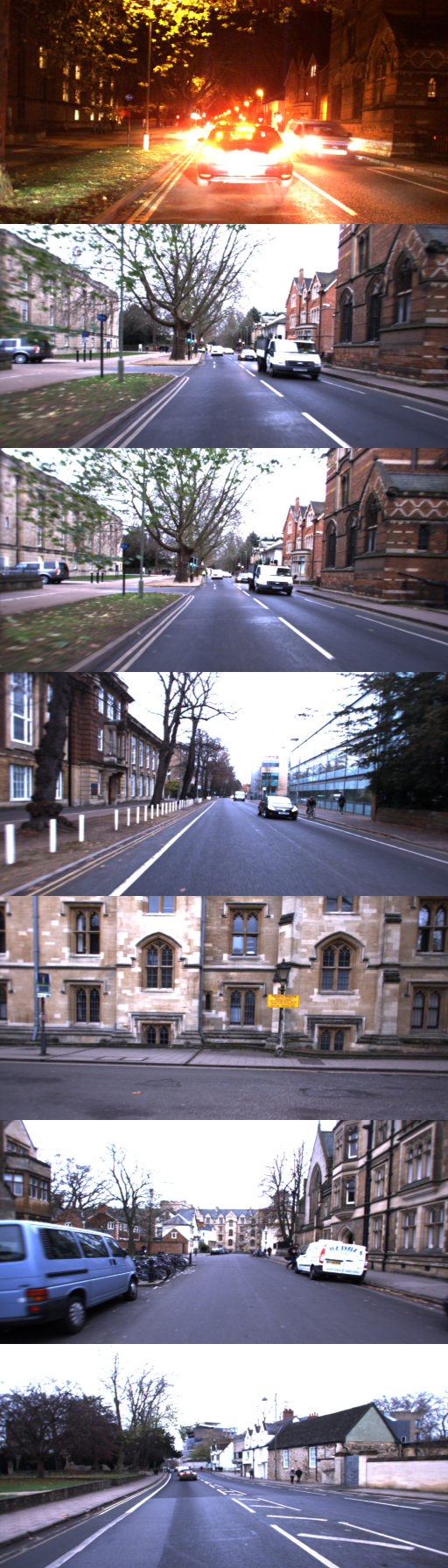}} \\ 
  GT \vspace{\vsOne} & & & & \\
  Ours-C5 \vspace{\vsOne} & & & & \\
  NV \vspace{\vsOne} & & & & \\
  Obj-C5 \vspace{\vsOne} & & & & \\
  Sem-C5 \vspace{\vsOne} & & & & \\
  NV-C5 \vspace{\vsOne} & & & & \\

\end{tabular}

\caption{\small Qualitative Results: For night time query images (top row), Ground Truth (GT) match (second row) and matches obtained from different methods are displayed (subsequent rows) including successful matches using our proposed representation (third row)}
\label{fig:vpr_qualRes}
\end{figure*}

%\newcommand{\widthOne}{0.12\textwidth}
%\newcommand{\scaleOne}{0.135}
%\newcommand{\shortP}{>{\centering\arraybackslash}p{\widthOne}}
%\begin{figure}
%    \centering
%\begin{tabular}{>{\centering\arraybackslash}p{\widthOne} >{\centering\arraybackslash}p{\widthOne} >{\centering\arraybackslash}p{\widthOne} >{\centering\arraybackslash}p{\widthOne} >{\centering\arraybackslash}p{\widthOne} >{\centering\arraybackslash}p{\widthOne} >{\centering\arraybackslash}p{\widthOne} } \\
%% p{\widthOne}p{\widthOne}p{\widthOne}p{\widthOne}p{\widthOne}}
%    \multicolumn{7}{c}{\includegraphics[scale=\scaleOne]{qualImages/qualRes_q110.png}} \\
%    \multicolumn{7}{c}{\includegraphics[scale=\scaleOne]{qualImages/qualRes_q372.png}} \\
%    \multicolumn{7}{c}{\includegraphics[scale=\scaleOne]{qualImages/qualRes_q558.png}} \\
%    \multicolumn{7}{c}{\includegraphics[scale=\scaleOne]{qualImages/qualRes_q608.png}} \\
%    Query & GT & Ours-C5 & NV & Obj-C5 & Sem-C5 & NV-C5 \\
%\end{tabular}

%\begin{figure}

\subsection{Study 2: Visual Place Recognition: Day versus Night}
The depth estimation network trained using our proposed approach is
able to learn appearance-robust features within the encoder. This is
particularly useful for visual place recognition under significant
appearance variations, for example, day versus night. The
state-of-the-art VPR methods use deep-learnt representations either
based on end-to-end
training~\cite{chen2017deep}\cite{radenovic2018fine}\cite{arandjelovic2016netvlad}
or indirectly derived from the internal layer
representations~\cite{garg19Semantic}
\cite{sunderhauf2015performance}\cite{anoosheh2019night}. For the
performance benchmark presented in this section, we directly compare
the convolutional features based image representations, extracted from
different networks. In this way, the repeatability of activation
patterns across day and night appearance conditions can be directly
evaluated.

Figure~\ref{fig:vpr_curve} shows the performance comparison among
different place representation methods. This includes flattened
$conv5$ representations from four different networks trained on
different tasks: \textit{Ours-C5} uses the encoder output from our
proposed network, trained to predict depth for night-time images;
\textit{NV-C5} uses VGG~\cite{simonyan2014very} based
NetVLAD~\cite{arandjelovic2016netvlad}, trained for place recognition;
\textit{Obj-C5} uses ResNet50~\cite{he2016deep}, trained for object
recognition; and \textit{Sem-C5} uses the encoder output of
RefineNet~\cite{lin2017refinenet} which is based on
ResNet101~\cite{he2016deep} and trained for dense semantic
segmentation. The latter has also been effectively used for
state-of-the-art place recognition descriptor
LoST~\cite{garg2018lost}. The flattened $conv5$ representations expect
a similar viewpoint between the compared pairs of images; for sake of
completion, we also include a viewpoint-invariant representation in
our comparisons: NetVLAD as \textit{NV} which uses $4096$-dimensional
descriptors. It can be observed that the feature representations based
on our depth encoder perform the best. While there is a significant
margin in performance for the flattened $conv5$ comparisons, the
proposed representation also outperforms the end-to-end learnt
viewpoint-invariant NetVLAD representation.

Figure~\ref{fig:vpr_qualRes} shows qualitative results for visual
place recognition under significant appearance variations. The first
row shows four query images captured under night time conditions;
their corresponding Ground Truth (GT) day-time image matches are shown
in the second row. In subsequent columns, image matches obtained
through different representation methods are displayed with the third
row comprising successful matches based on our proposed
representation. The incorrect matches using other methods in the first
column seem to indicate a bias in their selection based on the presence
of a vehicle in the query image. In the second row, it can be observed
that most of the retrieved matches comprise buildings viewed from far
with an oblique viewpoint, however, only the proposed representation
is able to obtain the correct match. We believe that learning to
predict depth per pixel for night time imagery enables the latent
representations to be more robust to perceptual aliasing caused by
appearance variations. Moreover, our proposed depth-estimation network
is trained in a completely unsupervised manner, where other
vision-based tasks like object recognition and semantic segmentation
would require labeled night-time data if they were to be used for
extracting appearance-invariant image representations for place
recognition.

\section{Conclusions and Future Scope} \label{sec:conc}
This paper discusses the problem of estimating depth from night-time
images, which suffers from poor visibility, non-uniform and
unpredictable variation in illumination arising from multiple and
possibly, moving light sources. The problem is tackled by applying a
patchGAN-based domain adaptation technique that allows an encoder to
adapt the \textit{features} obtained from the night-time images to acquire the
attributes of day-time features so that a decoder trained on day-time
images could be directly used for estimating depth from these adapted
features. The proposed novel approach is completely unsupervised as it does
not necessitate the availability of either explicit ground truth
signals (obtained from range sensors) or implicit supervision signals
obtained from multi-view (spatial / temporal) images. Unlike many of
the existing methods, the proposed method also does not require
generating simulated data which is considerably difficult for
night-time images. The efficacy of the proposed approach is
demonstrated through extensive analyses on the challenging Oxford
RobotCar dataset. Its usefulness is also demonstrated through its
application to a visual place recognition problem where the feature
representation obtained from our depth encoder is shown to outperform those obtained from 
the existing state-of-the-art methods. The proposed approach has
some limitations which will form the scope for future
investigations. As shown in Figure \ref{fig:failure_cases}, our method
can not deal with saturated regions (bright lights), very low
illuminated regions and thin structures, such as traffic signal poles. This could be solved
to some extent by incorporating semantic information in the learning
process.  Secondly, instead of learning two
separate encoders - one for day-time images and the other for
night-time images, it would be good to have one encoder model which
could be trained to learn context-specific features which are unique
to different styles rather than image-specific features. These
context-specific features could then provide the necessary semantics
to deal with the above failure cases.

\bibliographystyle{splncs04}
\bibliography{ref}
\end{document}